\pdfoutput=1
\documentclass[11pt]{article}

\usepackage{EMNLP2022}

\usepackage{times}
\usepackage{latexsym}

\usepackage[T1]{fontenc}
\usepackage[utf8]{inputenc}

\usepackage{microtype}

\usepackage{inconsolata}

\usepackage{graphics}
\usepackage{graphicx}
\usepackage{amsmath}
\usepackage{amsfonts}
\usepackage{mathbbol}
\usepackage{multirow}
\usepackage{array}
\usepackage{booktabs}
\usepackage{hyperref}

\title{ConstGCN: Constrained Transmission-based Graph Convolutional Networks for Document-level Relation Extraction}

\author{Ji Qi$^1$, Bin Xu$^1$, Kaisheng Zeng$^1$, Jinxin Liu$^1$, \\ \textbf{Jifan Yu}$^1$, \textbf{Qi Gao}$^2$, \textbf{Juanzi Li}$^1$, \textbf{Lei Hou}$^1$ \\
  $^1$Department of Computer Science and Technology, Tsinghua University \\
  $^2$Beijing Kedong Electric Control System Co. Ltd. \\
  \texttt{qj20@mails.tsinghua.edu.cn}, \texttt{xubin@tsinghua.edu.cn}
}

\begin{document}
\maketitle
\begin{abstract}
Document-level relation extraction with graph neural networks faces a fundamental graph construction gap between training and inference - the golden graph structure only available during training, which causes that most methods adopt heuristic or syntactic rules to construct a prior graph as a pseudo proxy.
In this paper, we propose \textbf{ConstGCN}, a novel graph convolutional network which performs knowledge-based information propagation between entities along with all specific relation spaces without any prior graph construction.
Specifically, it updates the entity representation by aggregating information from all other entities along with each relation space, thus modeling the relation-aware spatial information.
To control the information flow passing through the indeterminate relation spaces, we propose to constrain the propagation using transmitting scores learned from the Noise Contrastive Estimation between fact triples.
Experimental results show that our method outperforms the previous state-of-the-art (SOTA) approaches on the DocRE dataset.
The source code is publicly available at \href{https://github.com/THU-KEG/ConstGCN}{https://github.com/THU-KEG/ConstGCN}.
\end{abstract}

\section{Introduction}
Document-level relation extraction (DocRE) aims to extract heterogeneous relational graphs of form $\{\mathcal{G}=(\mathcal{E},\mathcal{R})\}$ in document, where the typed entities as nodes and multiple directional semantic relations as edges. In contrast to sentence-level RE \cite{qin2018robust,gao2020neural}, DocRE has been a growing interest by extracting relations beyond the sentence boundaries\cite{yao-etal-2019-docred} to ensure the information integrity.

\begin{figure}
    \centering
    \includegraphics[scale=0.34]{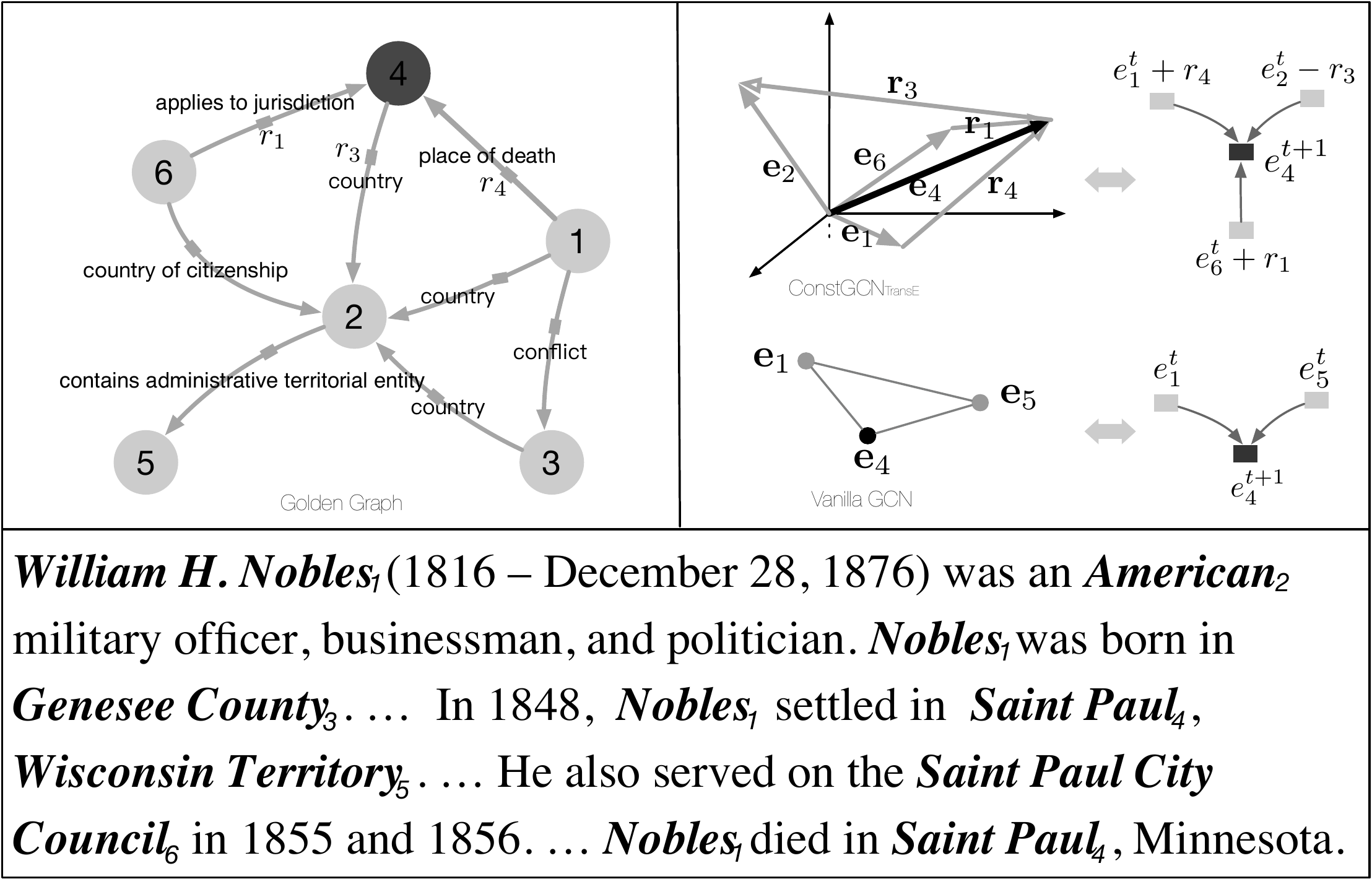}
    \caption{An example of DocRED document (bottom) with one of its golden multi-relational graphs (upper left). On the upper right, compared to the vanilla GNNs updating entity by accumulating representations of syntactically adjacent entities on the pseudo graph, the proposed ConstGNN models the relation-aware structural representation of entity  by performing knowledge-based information propagation.}
    \label{fig:example}
\end{figure}

Previous DocRE methods tend to apply the graph neural networks (GNNs) \cite{kipf2016semi,velivckovic2017graph} as the core component, and numerous variant of GNNs have proposed \cite{guo-etal-2019-attention,christopoulou2019connecting,zeng-etal-2020-double,xu2021document}. Similar to traditional GNNs modeled on the observable graph structures (e.g. social networks \cite{huang2019graph} and academic citation network \cite{feng2020graph}), these models all require a pre-specified graph construction.
They mainly either rely on the heuristic rules of intra(inter)-sentential information of entities and mentions \cite{zeng-etal-2020-double,christopoulou2019connecting}, or leverage the syntactic rules of dependency paths built by an external parser \cite{sahu-etal-2019-inter,guo-etal-2019-attention} to serve as the prior graph structure of GNNs.
We consider such graph structure as a \textit{pseudo graph structure}, for it establishes each edge between a pair of entities as a binary association based on the task-independent auxiliary information (heuristic/syntactic rules).

However, the golden edges that describe the relationships between two entities contain multi-type abundant semantics.
Thus, previous approaches suffer from two major intrinsic issues.
First, \textbf{Hindered Propagation Issue}: the construction of the pseudo graph structure ignores many actual relational edges between entities, which hinders the effective information acquisition and dissemination.
Second, \textbf{Noisy Representation Issue}: the simple information accumulation based on the binary associative edges on the pseudo graph makes them struggle to model relation-aware structural knowledge, which further results in noisy representations and harms the performance.
For example, in Figure \ref{fig:example}, the entity \textit{Saint Paul}'s representaion is updated by simply accumulating the representations of its syntactically adjacent entities, making it similar to entities \textit{Nobles} and \textit{Wisconsin Territory}, while they are completely different entities with relation connections \textit{place\_of\_birth}.

Instead of introducing the prior pseudo graph structures, we present \textbf{ConstGCN}, a novel \textbf{Cons}trained \textbf{T}ransmission-based \textbf{G}raph \textbf{C}onvolutional \textbf{N}etwork that performs knowledge-based information propagation between entities along with all relation spaces without any prior graph construction which explicitly models the semantics of various relationships.
Specifically, we innovatively propose the knowledge-based information propagation for DocRE by leveraging the flexible Knowledge Graph Embedding (KGE) approaches~\cite{wang2017knowledge} into a general transmitting operation. At each graph convolution step, it updates the entity representation by aggregating knowledge-based information broadcasted from neighbor entities along with all relation spaces.
Thus, entity representations containing the relation-aware structural semantics are learned effectively and directly.
Due to the agnostic nature of golden graph structure in documents, it is difficult to rigorously follow the relational paths to transmit information. We propose the transmitting scores to constrain the information flow through the indeterminate relational edges, where the scores are learned jointly from the Noise Contrastive Estimation (NCE) \cite{mikolov2013distributed}. It allows the model to learn the semantic representations while maintaining the original relation-aware structural information.
As shown in figure \ref{fig:example}, compared to the vanilla GNNs, our model learns the representations with an isomorphic structure to the golden heterogeneous graph in the document.

We conduct extensive experiments on DocRED, a large-scale human-annotated dataset including heterogeneous graphs among entities in each document. The results show that our model achieves the SOTA performance compared to previous methods.
In addition to the proposed graph convolutional network, we further demonstrated the compatibility of representation learning from documents and knowledge graphs. The contributions of our work are summarized as follows:

 \begin{itemize}
     \item We present a novel graph convolutional network, ConstGCN, that can naturally model the heterogeneous graph structure including indeterminate edges based on knowledge-based information propagation.
     \item We propose the approach of constrained transmission, which allows the model to learn the entity representations while maintaining the original relation-aware structural information.
     \item We conduct experiments on the DocRE task and achieve the SOTA performance. This work also demonstrates the compatibility of representation learning from documents and knowledge graphs.
 \end{itemize}

\section{Preliminary}

\subsection{Document-level Relation Extraction}

Given a textual document $\mathcal{D}=\{w_i\}_{i=1}^{|\mathcal{D}|}$ consisting of a sequence of words and a set of typed entities $\mathcal{E}=\{e_i\}_{i=1}^{|\mathcal{E}|}$, where each entity refers to a set of mentions $e=\{m_i\}_{i=1}^{|e|}$ which is a sequence of words in the document. The task of DocRE aims to extract the heterogeneous graphs $\{(e_i,r_k,e_j)|e_i,e_j\in\mathcal{E},r_k\in\mathcal{R}\}$, in which a pair of entities nodes $e_i,e_j$ may have multiple edges and each edge $r_k$ refers to a specific relation type, and $\mathcal{R}$ is the set of predefined relations.

\begin{figure*}[h]
    \centering
    \includegraphics[scale=0.255]{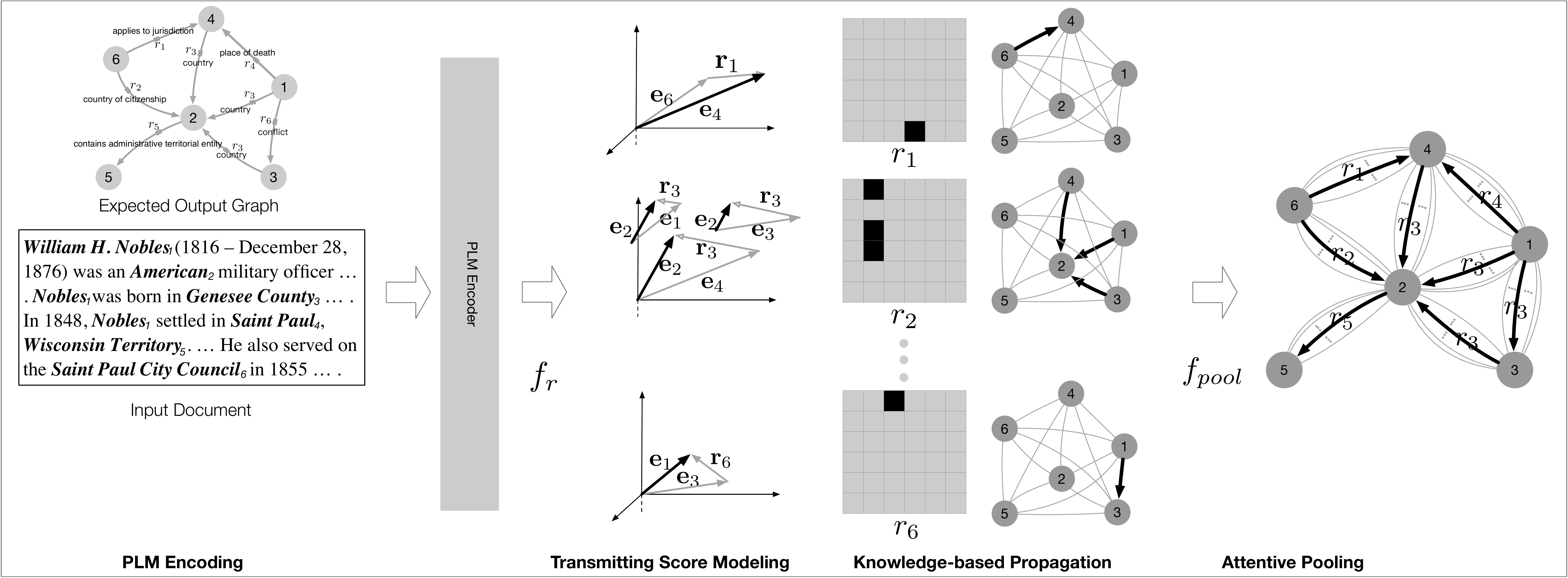}
    \caption{Overview of ConstGCN. Given the representations of entities, ConstGCN first computes the transmitting scores between entities in all relation spaces and then performs the graph convolution that updates each entity by transmitting its neighboring information along with projection spaces of all relations under the constraints of transmitting scores. Finally, the entity representations that model the structural semantics of heterogeneous graphs are learned and further used to predict the relational classes.}
    \label{fig:framework}
\end{figure*}

\subsection{Knowledge Graph Embedding}

Given a knowledge graph (KG) consisting of a collection of triples $\mathcal{G} = \{(e_i, r_k, e_j) | e_i,e_j\in\mathcal{E}, r_k\in\mathcal{R}\}$ with the sets of pre-defined types. The task of KGE aims to learn the vectorial representations $\mathbf{e}_i, \mathbf{r}_k, \mathbf{e}_j$ modeling the heterogeneous structural information based on a scoring function $d_r$.
Depending on the scoring function used, typical methods can be divided into two categories: the translational distance-based approaches e.g. (TransE\cite{bordes2013translating} and RotatE\cite{sun2019rotate}) and the semantic matching-based approaches (e.g. DistMult\cite{yang2014embedding} and ComplEx\cite{trouillon2016complex}). As shown in table \ref{tab:KGE_methods}, the key idea behind both categories of methods is to transmit the representation of head entity in relation-specific projection spaces to approximate tail entity. We thus define a unified transmitting operation $\oplus$ for the existing typical KGE approaches,

\begin{table}[t]
    \renewcommand{\arraystretch}{1.1}
    \centering
    \begin{tabular}{p{1.3cm} p{5.6cm}}
      \toprule[1.2pt]
        \textbf{Methods} &  \textbf{Scoring Function} \\
      \hline
        TransE & $d_r(e_i,r_k,e_j) = \gamma - ||\mathbf{e}_i + \mathbf{e}_j - \mathbf{r}_k||$ \\
    %   \hline
        RotatE & $d_r(e_i,r_k,e_j) = \gamma -||\langle\mathbf{e}_i \circ \mathbf{e}_j - \mathbf{r}_k\rangle||$ \\
    %   \hline
        DistMult & $d_r(e_i,r_k,e_j) = \langle\mathbf{e}_i, \mathbf{e}_j, \mathbf{r}_k\rangle$ \\
    %   \hline
        ComplEx & $d_r(e_i,r_k,e_j) = Re(\langle\mathbf{e}_i, \mathbf{e}_j, \mathbf{r}_k\rangle)$ \\
      \hline
    \end{tabular}
    \caption{The scoring functions for typical KGE models. The L1-norm is used for all distance based models and the subscript $||\cdot||_1$ is dropped for brevity.}
    \label{tab:KGE_methods}
\end{table}

\begin{align}
    \mathbf{e} \oplus \mathbf{r} = \begin{cases} \mathbf{e}+\mathbf{r}, \quad(\text{TransE}) \\
                    \langle\mathbf{e},\mathbf{r}\rangle, \quad(\text{DistMult}) \\
                    Re(\langle\mathbf{e},\mathbf{r}\rangle), \quad(\text{ComplEx}) \end{cases}
\end{align}
where $\langle\cdot\rangle$ denotes the generalized dot product, and $Re$ is the operation that returns the real part of a complex value.
The transmitting operation above will be used to perform knowledge-based message passing in document.

\section{Methodology}

In this section, we introduce the details of the proposed model. The overall framework is illustrated in figure \ref{fig:framework}. Based on the representations of entities obtained from a PLM encoder, ConstGCN is composed of $T$ graph  convolutional layers and each layer has two computational steps: the computation of transmitting scores and the computation of message passing under the constraints of transmitting scores.

\subsection{PLM Encoding}

Given an input document consisting of a sequence of words with a set of typed entities, we first insert a special token "*" at the start and end of entity mentions based on the entity marker tecnique \cite{zhang-etal-2017-position,shi2019simple,baldini-soares-etal-2019-matching}. For the processed document $\mathcal{D}=\{x_i\}_{i=1}^{|\mathcal{D}|}$,  we then employ a pre-trained language model (e.g. BERT \cite{devlin-etal-2019-bert}) to get contextual sequence representations:

\begin{equation}
\mathbf{H} = (\mathbf{x}_1, ..., \mathbf{x}_{|\mathcal{D}|})=\text{PLM}(x_1, ..., x_{|\mathcal{D}|}),
\end{equation}
where $\mathbf{x}_i\in\mathbb{R}^{d^w}$ refer to the contextualized embedding of $i$-th token.
For those documents that the sequence length are longer than the maximum input length of encoder, we compute the representations of overlapping tokens by averaging their embeddings from different windows. Then, we utilize the embedding of special token "*" at the start of $u$-th mention to represent the mention $\mathbf{m}_u$. For $i$-th entity with its mentions $e_i=\{m_u\}_{u=1}^{|e_i|}$, we compute an initial coarse-grained entity representation by averaging its coreference mentions with the logsumexp pooling function \cite{zhang2019ernie}:

\begin{equation}
\mathbf{e}_i = \log \sum_{m_u\in e_i} \exp(\mathbf{m}_u),
\end{equation}

we use these representations of entities obtained from the PLM encoder as the initialization of entity nodes  $\mathbf{E}^{(0)}=[\mathbf{e}_1,\mathbf{e}_2,..., \mathbf{e}_{|\mathcal{E}|}]^{\top}\in\mathbb{R}^{|\mathcal{E}|\times d}$.

\subsection{Constrained Transmission-based GCN}

To model the multi-relational graph structure, ConstGCN updates the entity representation by receiving information broadcasted from other entities along with all relation spaces. We thus introduce the representations of relations.
Instead of independently defining an embedding for each type of relation, we use a variant of the basis formulations\cite{schlichtkrull2018modeling} to linearly combine a set of basis vectors to promote generalization. Formally, let $\{\mathbf{z}_1, ..., \mathbf{z}_{\mathcal{B}} | \mathbf{z}_b\in\mathbb{R}^{d}\}$ be a set of basis vectors, a relation representation is given as:

\begin{align}
\mathbf{r}_k = \sum_{b=1}^{\mathcal{B}}\beta_{k_b}\mathbf{z}_b,
\end{align}
where $\beta_{k_b}\in\mathbb{R}$ is specific learnable scalar weight corresponding to $k$-th relation space.
We refer to these representations of relations as the initial representations of all relations $\mathbf{R}^{(0)}=[\mathbf{r}_1, ..., \mathbf{r}_{|\mathcal{R}|}]^{\top} \in\mathbb{R}^{|\mathcal{R}|\times d}$ for the subsequent constrained transmission-based graph convolution.

\paragraph{Modeling transmitting scores with NCE}
Due to the agnostic nature of relational edges, we need to broadcast the information of entities along with all relation spaces. Therefore, we need to measure the probability that there is a specific relationship between any two nodes to control the information flow.

Specifically, a simplified NCE objective \cite{gutmann2012noise,mikolov2013distributed} can be define as:

\begin{align}
\begin{split}
  \mathcal{L}_{nce} = &\log\sigma(\phi(\mathbf{v}_i,\mathbf{v}_j)) \\
    &+ \sum_{k}\mathbb{E}_{v_k'\sim p_n(v)}[\log\sigma(-\phi(\mathbf{v}_k', \mathbf{v}_i))],
\end{split}
\end{align}
where it differentiate positive sample pairs $(v_i, v_j)$ from noisy sample pairs $(v_k', v_i)$ drawing from the noise distribution $p_n$ by means of logistic regression, and $\phi$ is a specific measure function.
Inspired by the KGE ideas, we naturally use the logistic term of NCE by replacing the measure function $\phi$ as the KGE scoring functions to calculate the transmitting score from entity $e_i$ to entity $e_j$ through relation space $r_k$:

% \begin{small}
\begin{align}
f_{r}(\mathbf{e}^{(t)}_i, \mathbf{r}^{(t)}_k, \mathbf{e}^{(t)}_j) = \sigma(d_r(\mathbf{e}^{(t)}_i, \mathbf{r}^{(t)}_k, \mathbf{e}^{(t)}_j)),
\label{contras-obj}
\end{align}
% \end{small}
where the score indicates the probability that there is a relational edge $r_k$ from $e_i$ to $e_j$.

At $t+1$-th layer, by performing the computation of equation (\ref{contras-obj}) for all entity pairs at each relation space, we obtain the matrices of transmitting scores within all relation spaces

\begin{align}
    [\mathcal{A}^{(t)}]_{kij} = f_{r}(\mathbf{e}^{(t)}_i, \mathbf{r}^{(t)}_k, \mathbf{e}^{(t)}_j),
\end{align}
where $\mathcal{A}^{(t)}\in\mathbb{R}^{|\mathcal{R}|\times|\mathcal{E}|\times|\mathcal{E}|}$ refer to the tensor of transmitting scores corresponding to all relation spaces at $t$-th convolution layer.

Note that the equation (\ref{contras-obj}) can be used as an objective for KGE learning, and thus it can be used to optimize the representations of entities and relations with a specific KGE approach simultaneously.

\paragraph{Passing semantic information under Constraints}
For a heterogeneous graph consisting of entity nodes and relational edges in a document, an entity has multifaceted structural semantic information based on its neighbors of particular relations, and such multifaceted information is difficult to model by the hindered message passing on binary associative edges.
We propose to update each entity representation by transmitting the representations of its relation-specific neighbors along with the
relation spaces to enhance the multifaceted structural information.

Due to the agnostic nature of relation edges in the real world document, we further constrain the propagation with the transmitting scores learned from the previous step to maintain the original relation-aware structure. For each entity, the update of entity representation is defined as $\mathbf{e}^{(t+1)}_i=\sum_{k=1}^{|\mathcal{R}|}\sum_{j=1}^{|\mathcal{E}|}f_{r}(\mathbf{e}^{(t)}_j, \mathbf{r}^{(t)}_k, \mathbf{e}^{(t)}_i)(\mathbf{e}^{(t)}_j\oplus\mathbf{r}^{(t)}_k)$, where the operation $\oplus$ denote the transmitting operation contingent upon
specific KGE methods introduced above.
For the scale of all entities, we can rewrite this comprehensive transmission in an elegant way by using tensor multiplication, and promote computational efficiency:

\begin{align}
    &\mathbf{E}^{(t+1)}_k= \tilde{\mathcal{A}}^{(t)}\otimes(\mathbf{E}^{(t)}\oplus \mathbf{M}_k^{(t)}), \\
    &\mathbf{E}^{(t+1)} = f_{pool}(\mathbf{E}^{(t+1)}_1, \mathbf{E}^{(t+1)}_2, ..., \mathbf{E}^{(t+1)}_{|\mathcal{R}|}),
\end{align}
where $\mathbf{E}^{(t+1)}_k\in\mathbb{R}^{|\mathcal{E}|\times d}$ is the entity representations corresponding to the $k$-th relation space, $\tilde{\mathcal{A}}^{(t)}$ refers to the transposed transmitting scores that transpose the matrices consisting of the values of the second and third dimensions of $\mathcal{A}^{(t)}$, and $\mathbf{M}_k^{(t)}=[\mathbf{r}^{(t)}_k, ..., \mathbf{r}^{(t)}_k]^{\top} \in\mathbb{R}^{|\mathcal{E}|\times d}$ is the matrix composed of $|\mathcal{E}|$ identical vectors $\mathbf{r}^{(t)}_k$.
The operation $\otimes:(\mathbb{R}^{|\mathcal{R}|\times|\mathcal{E}|\times|\mathcal{E}|},\mathbb{R}^{|\mathcal{E}|\times d})\rightarrow\mathbb{R}^{|\mathcal{R}|\times|\mathcal{E}|\times d}$ refers to the tensor multiplication.
Similar to the pooling functions that aggregate feature maps along channel dimension in the computer vision, the function $f_{pool}(\mathcal{X}):\mathbb{R}^{d_1 \times d_2\times d_3}\rightarrow\mathbb{R}^{d_2\times d_3}$ aggregates the matrices comprised of the values with $d_2$ and $d_3$ dimensions of $\mathcal{X}$ along the $d_1$ dimension. There are multiple pooling operations can be used, such as $f_{pool}^{mean}$, $f_{pool}^{max}$ and $f_{pool}^{sum}$ for \textit{mean}, \textit{max}, \textit{sum}.

We introduce an attentive pooling function $f^{att}_{pool}$ that aggregates the representations of entities according to their importance in each relation space:

\begin{align}
\begin{split}
    &\mathbf{Z}^{(t)} = softmax(\frac{\mathbf{R}^{(t)}(\mathbf{E}^{(t)})^{\top}}{\sqrt{d}}),  \\
    &f_{pool}^{att}: \quad \sum_{k=1}^{|\mathcal{R}|}(\mathbf{z}_{k}^{(t)})^{\top}\cdot(\mathbf{E}^{(t+1)}_k),
\end{split}
\end{align}
where $\mathbf{z}_{k}^{(t)}\in\mathbb{R}^{|\mathcal{E}|}$ refers to the $k$-th row of $\mathbf{Z}^{(t)}$ indicating the importance of entities with respect to the relation space $r_k$.

By performing $T$ layers of the graph convolution, the representations of entities $\mathbf{E}^{(T)}$ and relations  $\mathbf{R}^{(T)}$ that modeled the multi-relational spatial information are obtained.

\subsection{Learning}

Based on the fine-grained entity representations obtained from ConstGCN, we further get the representation of each entity pair by supplementing the related local tokens information with the localized context pooling \cite{zhou2021document}. For each entity pair $(e_i, e_j)$ in current document, we get the localized representation by an attention-based weighted summation:

\begin{align}
\mathbf{c}^{(i,j)} = (\boldsymbol{\alpha}^{(i)}\cdot\boldsymbol{\alpha}^{(j)})\mathbf{H},
\end{align}
where $\boldsymbol{\alpha}^{(i)}\in\mathbb{R}^{|\mathcal{D}|}$ refer to the the averaged attention scores from $i$-th enttiy to all tokens for all layers in the pre-trained language model. Then the augmented representation is obtained by:

\begin{align}
  \begin{cases}
    \bar{\mathbf{e}}_i = \tanh(\mathbf{W}_s\mathbf{e}_i^{(T)} + \mathbf{W}_{c_1}\mathbf{c}^{(i,j)}), \\
    \bar{\mathbf{e}}_j = \tanh(\mathbf{W}_o\mathbf{e}_j^{(T)} + \mathbf{W}_{c_2}\mathbf{c}^{(i,j)}),
  \end{cases}
\end{align}
where $\mathbf{W}_s, \mathbf{W}_o, \mathbf{W}_{c_1}, \mathbf{W}_{c_1} \in\mathbb{R}^{d\times d}$ are the learnable weight matrices. Based on the derivation above, we employ two learning objective to optimize the model. The first is the cross-entropy-based classification objective to learn the final labels of relational classes between entities, and the second is a contrastive learning objective for the representation learning of entities and relations.

\subsubsection{Classification objective}

\begin{table*}
\begin{small}
\centering
\begin{tabular}{>{\centering\arraybackslash}m{2cm}l>{\centering\arraybackslash}m{1cm}>{\centering\arraybackslash}m{1cm}>{\centering\arraybackslash}m{1cm}>{\centering\arraybackslash}m{1cm}}
\toprule[1.5pt]
  \multirow{2}{1.2cm}{\textbf{Category}} & \multirow{2}{5cm}{\textbf{Model}} & \multicolumn{2}{c}{\textbf{Dev}} & \multicolumn{2}{c}{\textbf{Test}} \\
\cline{3-6}
  & & Ign F1 & F1 & Ign F1 & F1 \\
\hline
  \multirow{11}{2.2cm}{\textit{Sequence-based Models}} & CNN$^{*}$ \cite{yao-etal-2019-docred}  & 41.58 & 43.45 & 40.33 & 42.26   \\
  & LSTM$^*$ \cite{yao-etal-2019-docred} & 48.44 & 50.68 & 47.71 & 50.07  \\
  & BiLSTM$^*$ \cite{yao-etal-2019-docred} & 48.87 & 50.94 & 48.78 & 51.06  \\
  & ContexAware$^*$ \cite{yao-etal-2019-docred} & 48.94 & 51.09 & 48.40 & 50.07  \\
  & BERT$_{BASE}^*$ \cite{wang2019fine} & - & 54.16 & - & 53.20 \\
  & BERT-2Phase$_{BASE}^*$ \cite{wang2019fine} & - & 54.42 & - & 53.92 \\
  & HIN-BERT$_{BASE}^*$ \cite{tang2020hin} & 54.29 & 56.31 & 53.70 & 55.60  \\
  & CorefBERT$_{BASE}^*$ \cite{ye2020coreferential} & 55.32 & 57.51 & 54.54 & 56.96 \\
  & CorefRoBERTa$_{LARGE}^*$ \cite{ye2020coreferential} & 57.35 & 59.43 & 57.90 & 60.25 \\
  & MIUK(three-view)-BERT$_{BASE}^*$ \cite{li2021multi} & 58.27 & 60.11 & 58.05 & 59.99   \\
  & ATLOP-BERT$_{BASE}^*$ \cite{zhou2021document} & 59.22 & 61.09 & 59.31 & 61.30 \\
  & ATLOP-RoBERTa$_{LARGE}^*$ \cite{zhou2021document} & 61.32 & 63.18 & 61.39 & 63.40 \\
\hline
  \multirow{11}{2cm}{\textit{Graph-based Models }} & GAT$^\dagger$ \cite{velivckovic2017graph} & 45.17 & 51.44 & 47.36 & 49.51  \\
  & AGGCN$^\dagger$ \cite{guo-etal-2019-attention} & 46.29 & 52.47 & 48.89 & 51.45  \\
  & EoG$^\dagger$ \cite{christopoulou2019connecting} & 45.94 & 52.15 & 49.48 & 51.82  \\
  & GCNN$^\dagger$ \cite{sahu-etal-2019-inter} & 46.22 & 51.52 & 49.59 & 51.62   \\
  & LSR-BERT$_{BASE}^*$ \cite{nan-etal-2020-reasoning} & 52.43 & 59.00 & 56.97 & 59.05  \\
  & GAIN-BERT$_{BASE}^\natural$ \cite{zeng-etal-2020-double} & 57.75 & 60.03 & 57.98 & 60.42  \\
  & HeterGSAN-BERT$_{BASE}^*$ \cite{xu2021document} & 58.13 & 60.18 & 57.12 & 59.45   \\
\cline{2-6}
  & \textbf{ConstGCN-BERT$_{BASE}$} & \textbf{59.32} & \textbf{61.27} & \textbf{59.58} & \textbf{61.55}  \\
  & \textbf{ConstGCN-RoBERTa$_{LARGE}$} & \textbf{62.01} & \textbf{63.91} & \textbf{62.04} &  \textbf{64.00} \\
\hline
\end{tabular}
\caption{Performance on the dev set and the test set of DocRED. The bottom two rows show results of our models with \textit{TransE} implementation. Results with $*$ are reported in their original papers. Results with $^\dagger$ are implemented in [Nan et al., 2020]. Results with $^\natural$ are our reproduced performance. Bold results indicate the optimal performances.}
\label{table:main_results}
\end{small}
\end{table*}

The probabilities of final relational classes for each pair of entities in the document are calculated as following:

\vspace{-12pt}
\begin{align}
    P(r|e_i,e_j) =\sigma(\bar{\mathbf{e}}_i \mathbf{W}_r \bar{\mathbf{e}}_j + b_r),
\end{align}
where $\mathbf{W}_r\in\mathbb{R}^{d\times d}, b_r\in\mathbb{R}$ are learnable model parameters. We utilize the adaptive thresholding method \cite{zhou2021document} that involves a learnable threshold class TH to differentiate the positive and negative classes. Then the cross-entropy based classification objective is defined as:

\begin{align}
\small
\begin{split}
  &\mathcal{L}_{cls_1} = -\sum_{r_k\in\mathcal{T}_P}\log(\frac{\exp(P(r_k|e_i,e_j))}{\sum_{r_k'\in\mathcal{T}_P\cup\{TH\}}\exp(P(r_k'|e_i,e_j))}), \\
  &\mathcal{L}_{cls_2} = -\log(\frac{\exp(P(TH|e_i,e_j))}{\sum_{r_k'\in\mathcal{T}_N\cup\{TH\}}\exp(P(r_k'|e_i,e_j))}), \\
  &\mathcal{L}_{cls} = \mathcal{L}_{cls_1} + \mathcal{L}_{cls_2},
  \nonumber
\end{split}
\end{align}
where $\mathcal{T}_P$ and $\mathcal{T}_N$ are the all positive relational triples set and all negative relational triples set respectively. At the test time, the positive relation classes are returned if the probabilities are higher than the TH classes.

\subsubsection{Contrastive learning objective} We optimize the transmission learning with the oise Contrastive Estimation (NCE)\cite{gutmann2012noise,mikolov2013distributed}. Instead of using a uniform negative sampling, we leverage the self-adversarial negative sampling method introduced by\cite{sun2019rotate} to alleviate the non-meaningful information problem,

\begin{small}
\begin{align}
 \begin{split}
     \mathcal{L}_{nce} = & -\sum_{i\neq j}\sum_{(e_i,e_j,r_k)\in\mathcal{T}_P}(\log\sigma(d_r(\bar{\mathbf{e}}_i, \bar{\mathbf{r}}_k, \bar{\mathbf{e}}_j)) \\
     &+ \sum_{(e_{q}',r_k,e_{q}'')\in\mathcal{T}_N^{q}}\varphi_{q}^k\log\sigma(-d_r(\bar{\mathbf{e}}_{q}', \bar{\mathbf{r}}_k, \bar{\mathbf{e}}_{q}''))), \\
   \varphi_{q}^k = &\frac{\exp\tau \sigma(-d_r(\bar{\mathbf{e}}_{q}', \bar{\mathbf{r}}_k, \bar{\mathbf{e}}_{q}''))}{\sum_{(e_{l}',r_k,e_{l}'')\in\mathcal{T}_N^{l}}\exp\tau \sigma(-d_r(\bar{\mathbf{e}}_{l}', \bar{\mathbf{r}}_k, \bar{\mathbf{e}}_{l}''))},
   \nonumber
\end{split}
\end{align}
\end{small}

where $\mathcal{T}_N^q$ refer to the negative triples sampled from current document $q$, and $\tau$ is the sampling temperature. Finally, the overall objective is a linear combination of these two objectives,

\begin{align}
\mathcal{L} = \mathcal{L}_{cls} + \mu\mathcal{L}_{nce},
\end{align}
where $\mu$ is the hyperparameter to balance these two different objectives.

\section{Experiments}

\subsection{Experimental Settings}

\paragraph{Dataset.} The proposed methods on a widely-used human-annotated dataset, DocRED \cite{yao-etal-2019-docred}, which is built based on Wikipedia and Wikidata, with 5,053 documents, 132,375 entities, 56,354 relational facts, and 96 relation types. More than 40.7\% relations can only be extracted across sentences, and 61.1\% relations require inference skills such as logical inference. We follow the standard split of the dataset, i.e., 3,053 documents for training, 1,000 for development and 1,000 for test.

\paragraph{Baselines.} We compare ConstGCN with sequential methods that adopt sequential encoders as the main architectures, including \textbf{CNN/LSTM}\cite{yao-etal-2019-docred}, \textbf{BERT-2Phase}~\cite{wang2019fine}, \textbf{HIN-GloVe/HIN-BERT}~\cite{tang2020hin}, \textbf{CorefBERT} \cite{ye2020coreferential}, \textbf{MIUK}~\cite{li2021multi} and \textbf{ATLOP}~\cite{zhou2021document}, and graphical methods leveraging GNNs as the backbone, including \textbf{GCNN} \cite{sahu-etal-2019-inter}, \textbf{GAT} \cite{velivckovic2017graph}, \textbf{AGGCN} \cite{guo-etal-2019-attention}, \textbf{EoG}~\cite{christopoulou2019connecting}, \textbf{GAIN}~\cite{zeng-etal-2020-double}, \textbf{LSR}~\cite{nan-etal-2020-reasoning} and \textbf{HeterGSAN}~\cite{xu2021document}.

\paragraph{Implementation.} For the foundational components, we use cased BERT$_{base}$ or RoBERTa$_{large}$ \cite{liu2019roberta} as the encoder to validate the efficiency for all BERT-based methods. For ConstGCN, we empirically set the number of relation basis vectors $\mathcal{B}=56$, and the pooling function as \textit{att} tuned on the dev set.
We investigate three implementations with different transmitting operations, \textbf{ConstGCN(TransE)}, \textbf{ConstGCN(DistMult)} and \textbf{ConstGCN(ComplEx)}, where the fixed margin of ConstGCN(TransE) is set to $\gamma=20.0$.
The number of graph convolutional layers are set to 2, 2, and 1 for the three implementations, respectively, and the negative sampling proportion in transmitting score's learning is set to $1/|\mathcal{T}_N^q|=1/40$ consistently for all implementations.
All hyperparameters are tuned based on grid search with the performance on dev set, and more training details are available in Appendix \ref{lbl:hyperparameter_settings}.

\begin{figure*}[h]
    \centering
    \includegraphics[scale=0.24]{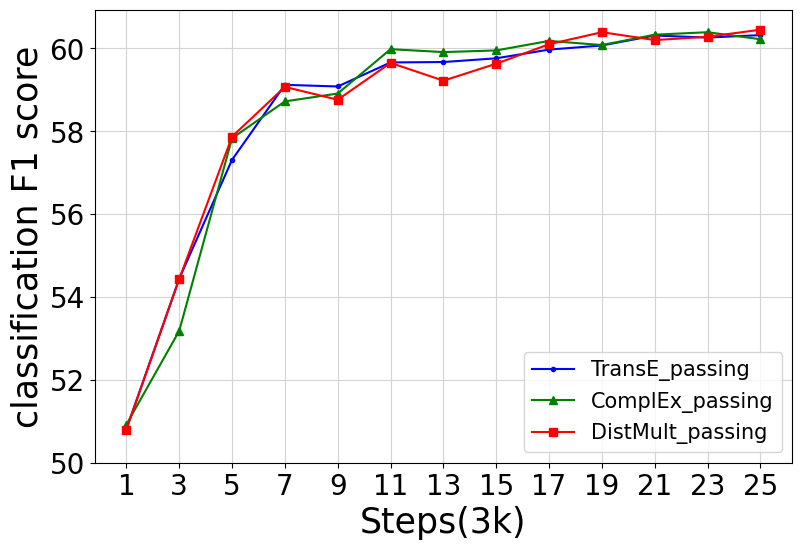}
    \includegraphics[scale=0.24]{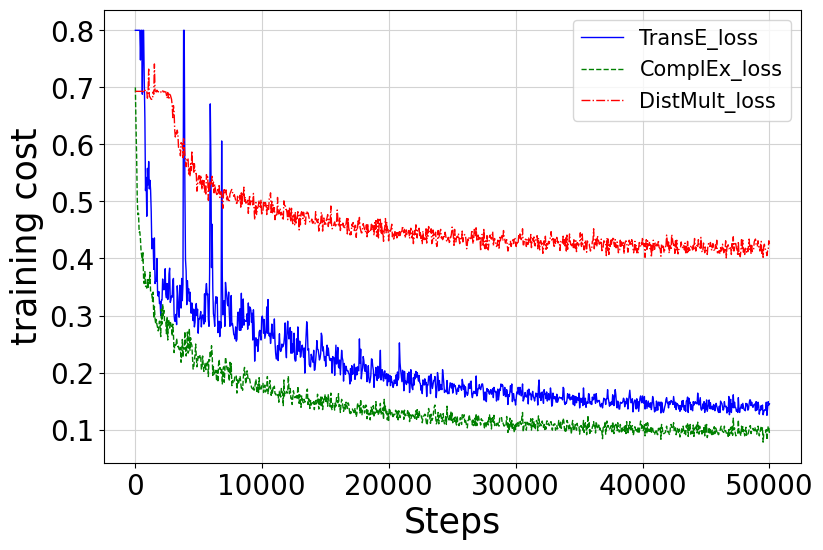}
    \includegraphics[scale=0.22]{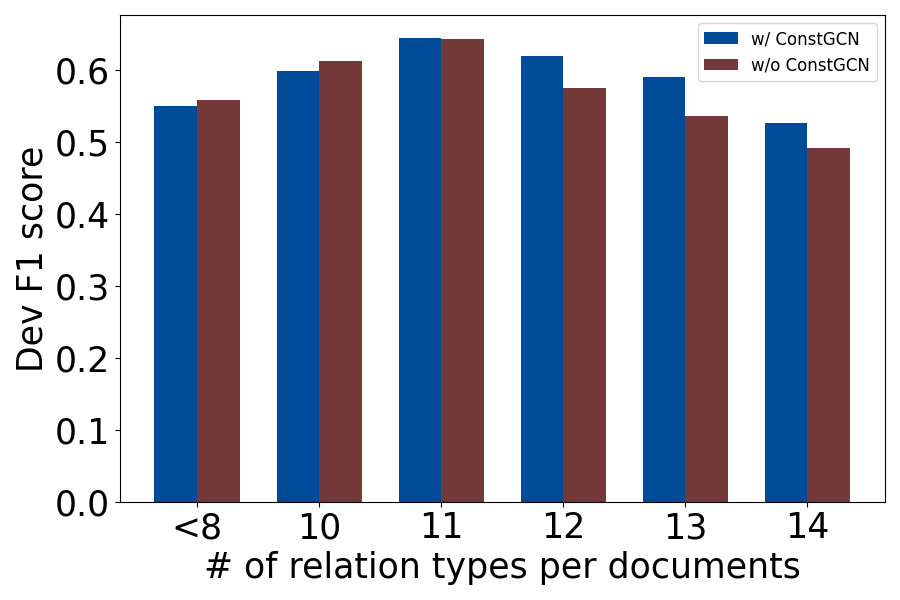}
    \caption{The first two figures show the dev F1 score with the transmitting costs, where the F1 score strictly increases as the transmission cost decreases until convergence. The third shows the dev F1 score with the different number of relation types on DocRED. ConstGCN shows more superiority when the number of relation types grows.}
    \label{fig:analysis_F1_kge}
\end{figure*}

\subsection{Main Results}

The main results on the DocRED dataset with the standard metrics F1 and Ign F1 are presented in Table \ref{table:main_results}, where Ign F1 calculate F1 scores excluding the common relation facts shared in the training and dev/test sets.
We group the existing models into 2 categories: (1) the \textit{Sequence-based Models} that adopt sequential encoder to encode and obtain the representations; (2) the \textit{Graph-based Models} that focus on modeling the heterogeneous graph structure with GNNs independent of encoder.

As shown in the Table \ref{table:main_results}, our methods achieve the state-of-the-art performance on all metrics and all splits.
Compared to the group of \textit{Graph-based Models}, our method outperforms the state-of-the-art GNNs-based model by 1.09\% F1 score and 2.1\% F1 score on the development set and test set respectively.
These results suggest that ConstGCN naturally learns more fine-grained
informative representations of entities during the constrained transmission-based graph convolution.
By using RoBERTa$_{LARGE}$ as the encoder, our ConstGCN further achieves the F1 scores of 63.91\% and 64.00\% for the both evaluation set, which are a new state-of-the-art results on DocRED.

\subsection{Model Analysis}

In this section, we further investigate the effectiveness of the transmission learning and the impact of ConstGCN on documents of different complexity.

\paragraph{How does the transmission learning affect the performance ?}

Figure \ref{fig:analysis_F1_kge} shows the learning curves of classification performance and transmission costs for all three implementations on the development set during the training.
As seen, the $F1$ scores strictly improve with the decreasing of transmitting costs for all three models, and the model achieves the optimal classification performances when the transmission costs converge. This shows that the informative entity representations learned by optimizing the constrained transmission-based graph convolution can lead to significant improvements in classification performance.  We also notice that the model using ComplEx transmitting operation performs the best in terms of convergence speed and optimal values compared to the other models, and it suggests that the model learns effective representations especially in the complex space. The visualization of the transmitting scores learned between entities in a specific document are shown in Appendix \ref{sec:visualization}.

\paragraph{What kind of environment is ConstGCN more effective for ?}

We further investigate the effectiveness of ConstGCN on documents of different relation-aware complexities to demonstrate its applicability.
As the third sub-figure~\ref{fig:analysis_F1_kge} shows, we split the development set of DocRED into 6 disjoint subsets by the number of relations types, and evaluate models trained with or without ConstGCN on each subset. Overall for both models, we found that their F1 performances tend to decrease when the number of relation types keeps growing.
However, when the number of relation types is larger than 10, the model w/ ConstGCN consistently exhibit significant better performance compared to the model w/o ConstGCN. This result indicates that the broadcasting of knowledge-based transmission in all relation spaces can learn relation-aware structural information and is more applicable to complex environment. Further, we can assume that when the complexity of the document exceeds that of DocRED, we can still get good performance by adjusting the transmitting operation in ConstGCN.

\begin{table}[hbt]
\begin{small}
    \renewcommand{\arraystretch}{1}
    \centering
    \begin{tabular}{l>{\centering\arraybackslash}m{2cm}c>{\centering\arraybackslash}m{2cm}c}
      \toprule[1.5pt]
        \textbf{Model} & \textbf{F1} & \textbf{AUC} \\
      \hline
        \textbf{ConstGCN (TransE)}     & 61.27 & 61.32 \\
         \quad \#layers $T=1$     & 47.52 & 44.46 \\
         \quad \#layers $T=3$     & 60.87 & 61.69 \\
         \quad $f_{pool}$ = \textit{mean}    & 61.23 & 61.32 \\
         \quad $f_{pool}$ = \textit{max}    & 61.20 & 61.40 \\
      \hline
        \textbf{ConstGCN (DistMult)}     & 61.06 & 62.21 \\
         \quad \#layers $T=1$     & 61.06 & 60.99 \\
         \quad \#layers $T=3$     & 61.48 & 61.50 \\
      \hline
        \textbf{ConstGCN (ComplEx)}     & 61.41 & 61.73  \\
         \quad \#layers $T=2$     & 60.89 & 62.39 \\
         \quad \#layers $T=3$     & 59.85 & 61.80 \\
      \hline
    \end{tabular}
    \caption{Ablation studies of ConstGCN-BERT$_{BASE}$. We change different implementations of components one at a time. These ablation results show that, the 2-layer ConstGCN can learn effective representations and benefit classification performance.}
    \label{tab:ablation_study}
\end{small}
\end{table}

\subsection{Ablation Study}
In this subsection, we examine the contributions of main components under different implementations.

Specifically, we explore the effectiveness of the number of graph convolution layers with different transmitting operations, and the performances of using different pooling functions. Table \ref{tab:ablation_study} shows the results on the development set.

Overall, we observe that a 2-layer transmission-based graph convolution learns effective entity representations and obtains the best classification F1 performance for all three models. In particular, the performance of ConstGCN (TransE) model deteriorates significantly when the number of layers of graph convolution was reduced to 1, while the other models maintain competitive performance with 1 layer. This suggest that the translation-based approaches require more transmitting steps to accumulate effective information than the semantic-based methods. Moreover, the model demonstrates consistent performance with three different pooling functions, \textit{att}, \textit{mean} and \textit{max}.

\section{Related Work}

\paragraph{Relation Extraction}
Early research efforts on relation extraction concentrate on extracting relations between entity pairs within a sentence.  Various approaches including feature-based methods \cite{mintz2009distant}, kernel-based methods \cite{bunescu2005shortest}, and deep neural networks-based methods \cite{qin2018robust,gao2020neural} have been shown effective in handling this problem.
However, due to the specificity of data structures in biomedical domain, some recent researches construct the document graph with heuristics and syntactical rules \cite{quirk2016distant,guo-etal-2019-attention}, and then perform inference to extract the binary interactions between biomedical entities (e.g. 3-ary relation between drug, gene, and mutation) in the entire document.

However, due to the specificity of data structures in biomedical domain, some recent researches construct the document graph with heuristics and syntactical dependencies \cite{quirk2016distant,guo-etal-2019-attention}, and then perform inference to extract the binary interactions between biomedical entities (e.g. 3-ary relation between drug, gene, and mutation) in the entire document.
More recently, with the large-scale general-purpose DocRE dataset proposed by \cite{yao-etal-2019-docred}, there has been a growing interest in extracting relations on such multi-mention multi-label environment \cite{wang2019fine,ye2020coreferential,tang2020hin,nan-etal-2020-reasoning,zeng-etal-2020-double,xu2021document,li2021multi}.

\paragraph{Knowledge Graph Embedding} The Knowledge Graph Embedding (KGE) \cite{wang2017knowledge} approaches aims to model semantic representations of entities and relations in the multi-relational graph structure properly, and it is extensively studied recently. Most of KGE approaches define a score function that models the entity and relation embeddings to constrain the valid triples with a higher score than the invalid ones. The KGE methods can be briefly classified into two major categories based on the type of scoring functions.
The translation-based methods including TransE\cite{bordes2013translating} and RotatE\cite{sun2019rotate} measure the socre as translating distance from head entity to tail enity along the relation space, and the semantic-based methods such as DistMult\cite{yang2014embedding} and ComplEx\cite{trouillon2016complex} exploit the score as semantic similarity between head and tail entities with the relation-specific projection space.

\section{Conclusion}

This paper proposes a novel graph convolutional network that naturally learns the relation-aware entity representations on heterogeneous graphs with indeterminate edges. To avoid the prior pseudo graph structure, we transmit entity representations along with all relation spaces to update each entity, in which the transmission is constrained with the transmitting scores learned from Noise Contrastive Estimation to maintain the original spatial information. Experimental results show that our method greatly advantages the DocRE task.
\clearpage

\section*{Limitation}
In this paper, we explore 3 typical KGE ideas into the knowledge-based graph convolution and prove their efficiency. However, there are many other approaches, such as KGE in hyperbolic space, which we do not validate in this paper. In the training of ConstGCN, it is a very complicated problem to perform better negative sampling of the entity-relation triples in the documents. In this paper, we follow the strong-weak negative sample strategy used in many KGE approaches, with a carefully chosen sampling ratio in the document, but this is still not sufficient. We have theoretically proven the compatibility of representation learning between natural texts and KGs based the ConstGCN, rather than verifying it on a case-by-case basis.

\bibliography{emnlp2022}
\bibliographystyle{acl_natbib}

\clearpage

\appendix

\section{Hyperparameter settings and implementation details}
\label{lbl:hyperparameter_settings}

We train ConstGCN on one NVIDIA RTX 3090 for a maximum of 30 epochs, it takes about 5 hours to finish training.
We tune the optimal hyperparameters using grid search based on the F1 score on dev set.
We set the dropout \cite{srivastava2014dropout} with rate 0.1, and clip the gradients to a max norm of 1.0. The optimization method is set to AdamW \cite{loshchilov2017decoupled} with an initial learning rate $1e^{-5}$ and a linear warmup\cite{goyal2017accurate} for the first 6\% steps with exponential decay.
As shown in Table~\ref{tab:setting_base_network} and \ref{tab:setting_ConstGCN}, the value of htperparameters we finally adopted are in bold.

\begin{table}[h]
    \centering
    \begin{tabular}{lr}
      \toprule[1.5pt]
        \textbf{Hyperparameter} & Value \\
      \hline
        Batch Size &  8, 6, \textbf{2} \\
        Learning Rate & \textbf{0.001} \\
        Activation Function & \textbf{ReLU}, Tanh \\
        Positive v.s. Negative Ratio & 1.0, 0.5, \textbf{0.25} \\
        Entity Type Embedding Size & \textbf{20} \\
        Coreference Embedding Size & \textbf{20} \\
        Dropout & 0.2, \textbf{0.6}, 0.8 \\
        Weight Decay & \textbf{0.0001} \\
      \hline
    \end{tabular}
    \caption{Settings for base network}
    \label{tab:setting_base_network}
\end{table}

\begin{table}[h]
    \centering
    \begin{tabular}{p{2cm}r}
      \toprule[1.5pt]
        \textbf{Hyperparameter} & Value \\
      \hline
        $T$ & \textbf{1}, \textbf{2}, \textbf{3} \\
        $f_{pool}$ & Max, Mean, \textbf{Att} \\
        $f_{r}$ & \textbf{TransE}, \textbf{ComplEx}, \textbf{DistMult} \\
        $\gamma$ & 4, 8, 12, 16, \textbf{20}, 24, 28 \\
        $\mathcal{B}$ & 48, \textbf{56}, 64, 72, 80, 88, 96\\
        $|\mathcal{N}_e^q|$ & 10, 20, \textbf{40} \\
        % $\epsilon$ & 1.0 \\
        $\tau$ & \textbf{1.0}, 2.0 \\
        $\mu$ & \textbf{0.001}, 0.01 \\
      \hline
    \end{tabular}
    \caption{Settings for ConstGCN}
    \label{tab:setting_ConstGCN}
\end{table}

\section{Visualization of Transmitting Scores}
\label{sec:visualization}

\begin{figure*}
    \centering
    \includegraphics[scale=0.15]{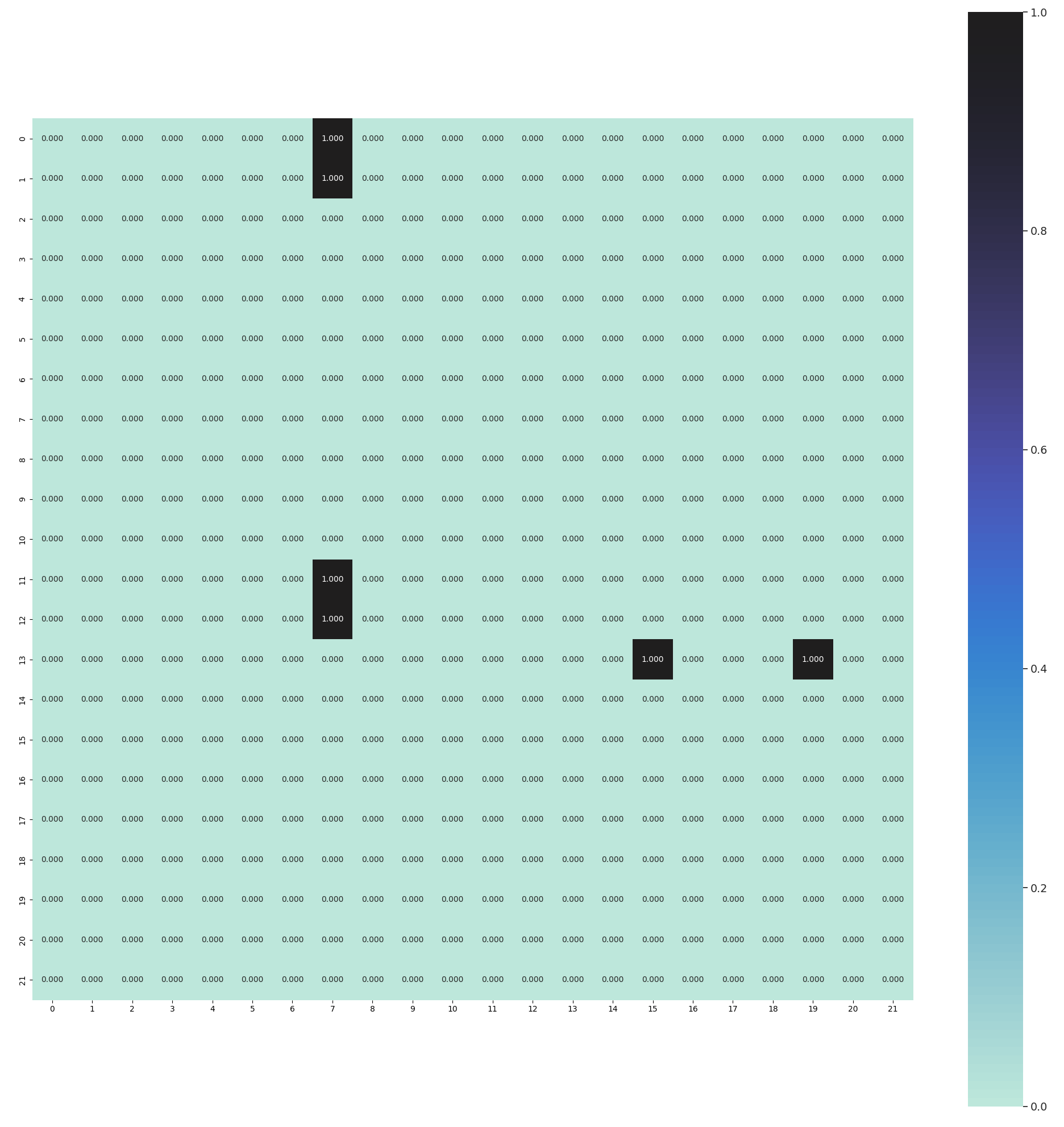}
    \includegraphics[scale=0.15]{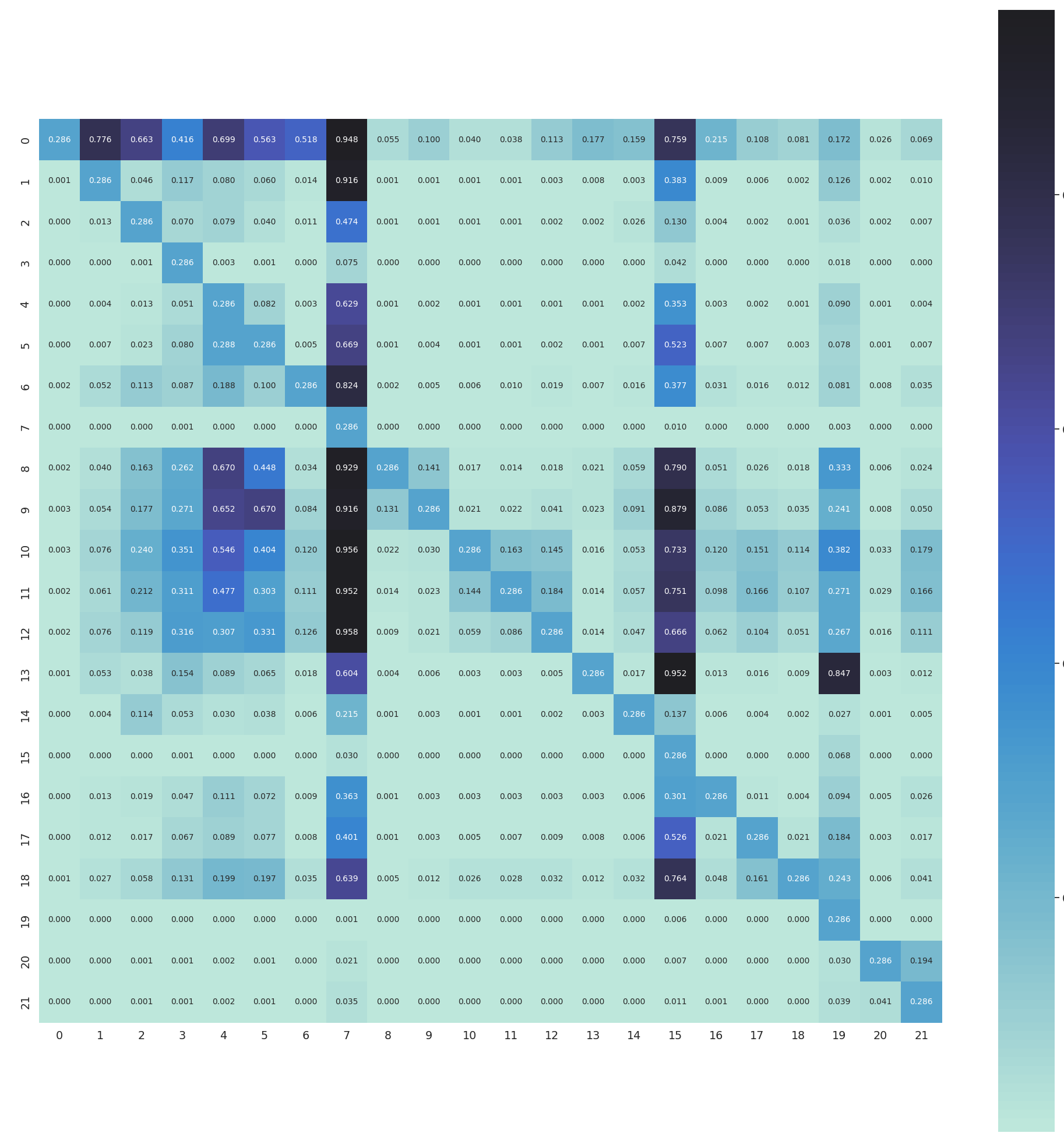}
    \caption{Visualization of the transmitting scores learned between all entities in the specific relation space of \textit{country}. Left: the golden adjacency matrix that each element represent a golden relation if the value is equal to 1; Right: the transmitting scores learned with the transmitting operation \textit{TransE}.}
    \label{fig:visulize_gold_transe}
\end{figure*}

\begin{figure*}
    \centering
    \includegraphics[scale=0.15]{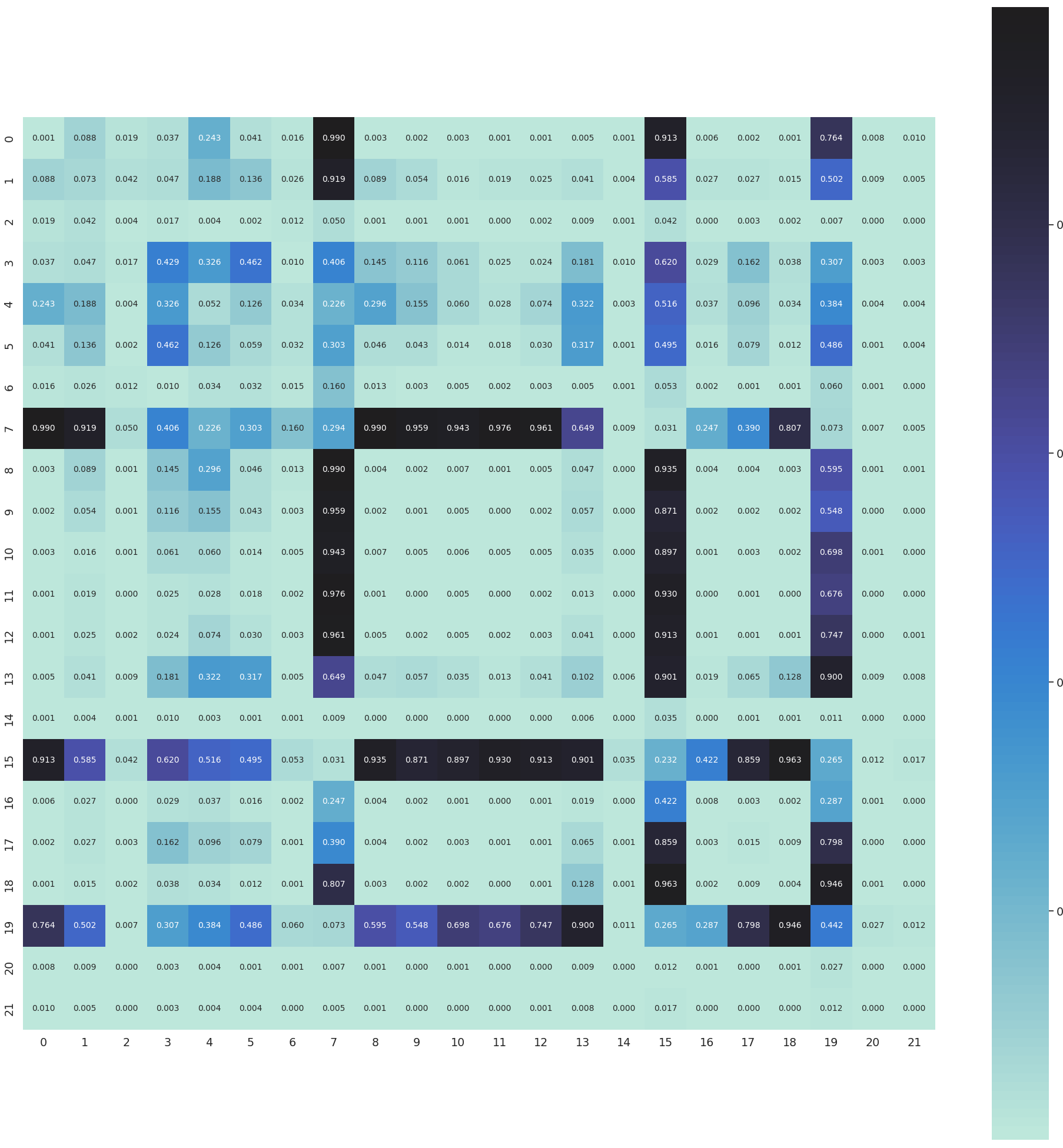}
    \includegraphics[scale=0.15]{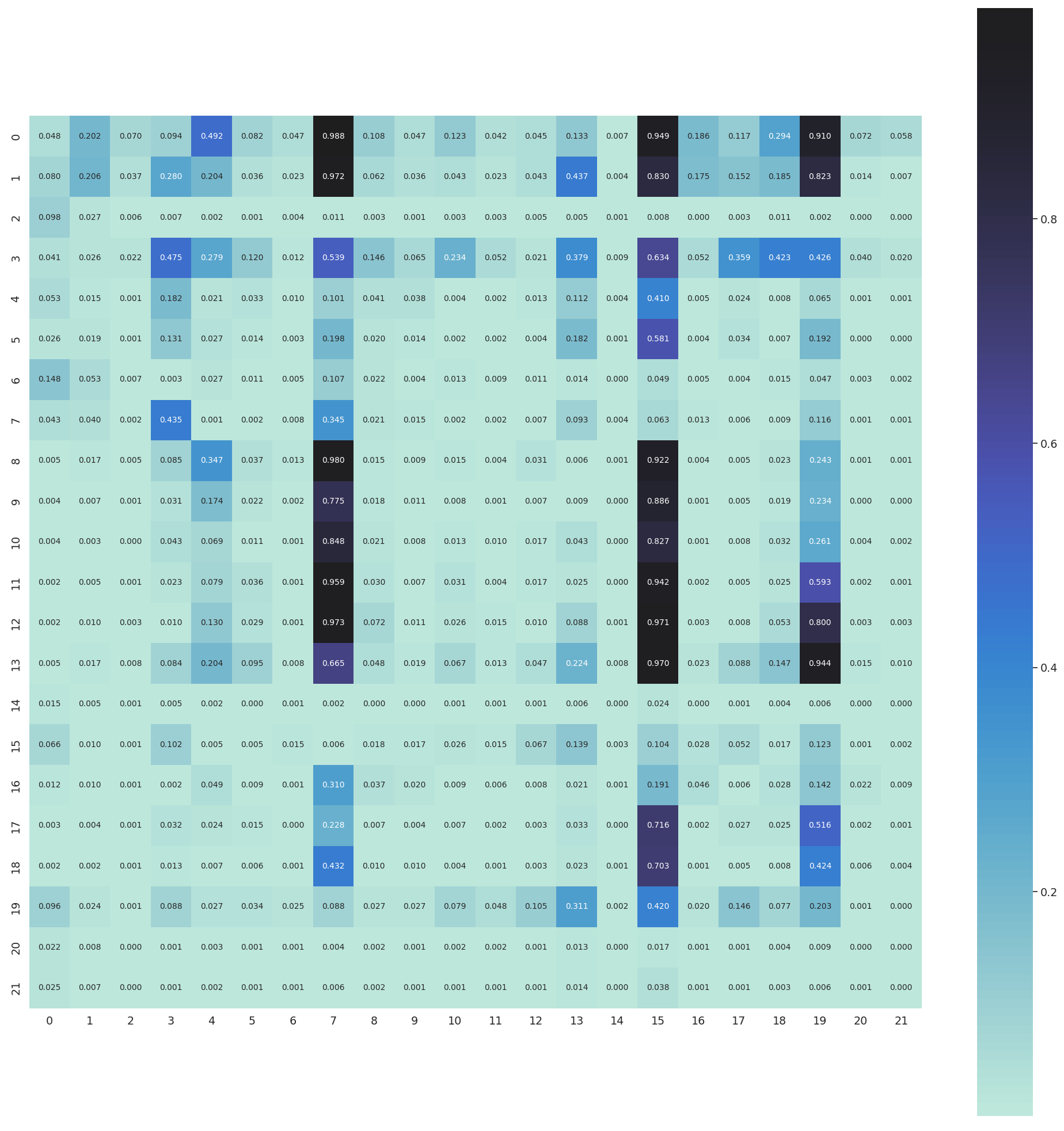}
    \caption{Visualization of the transmitting scores learned between all entities in the specific relation space of \textit{country}. Left: the transmitting scores learned with the transmitting operation \textit{DistMult}; Right: the transmitting scores learned with the transmitting operation \textit{ComplEx}.}
    \label{fig:visulize_distmult_complex}
\end{figure*}

We visualize the transmitting scores between all entities in a specific relation space, \textit{country}, in a document from the development set, which are predicted by the optimal models of three implementations. The values are shown in Figure \ref{fig:visulize_gold_transe} and Figure \ref{fig:visulize_distmult_complex}.  We find that the transmitting scores calculated during the graph convolution consistently conform with the adjacency matrix of golden graph, suggesting that our models learn relation-aware structural information for entities effectively.
Besides, based on the visualized figure, we can clearly
find that the problem of N-1 that has been identified in traditional TransE still exists in ConstGCN, and it inspires us for future improvements.

\section{Error Analysis}
\label{sec:error-analysis}

\begin{figure}[hbt]
    \centering
    \includegraphics[scale=0.34]{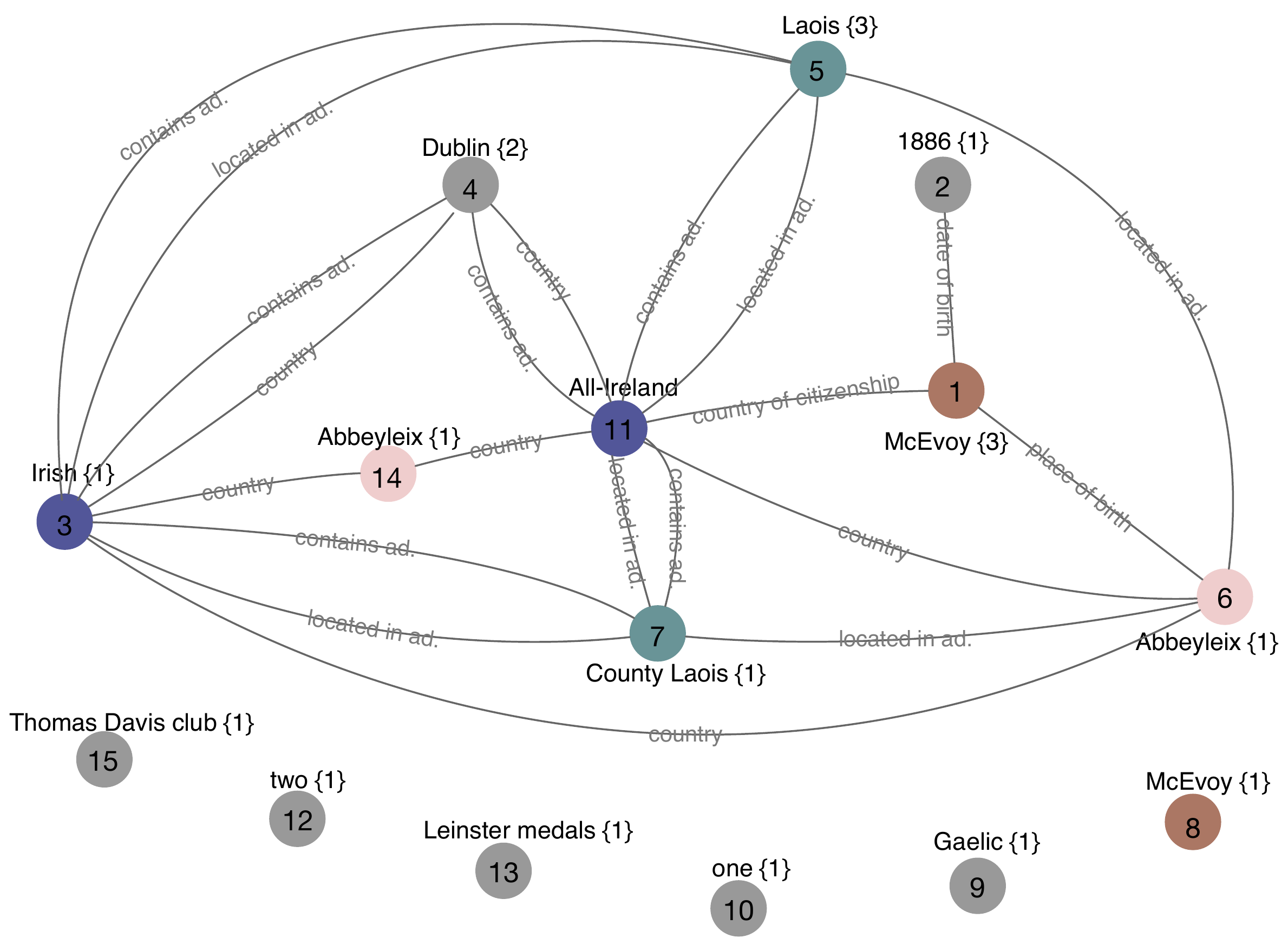}
    \caption{An example of input document and annotated golden relational graphs from DocRED. The \textit{contains ad.} and \textit{located in ad.} refer to the relations \textit{contains administrative territorial entity} and \textit{located in the administrative territorial entity}, respectively.The number inside the brackets after the entity name indicates the number of coreferences the entity have.}
    \label{fig:append_eg1}
\end{figure}

An annotated example that our model is underperforming on it is shown in bellow. We can see a frustrating scene: there are a large number of mistakes and omissions in the labeled data, especially since many entities that actually co-referenced are annotated as separate entities. In the Figure~\ref{fig:append_eg1}, entities of the same color, except for gray, indicate that they should be labeld as co-referencing entities. However, these omitted annotations make it difficult for model to accurately distinguish semantic boundaries.

\begin{small}
    \textit{Edward P. "Ned" McEvoy (born 1886) was an Irish hurler who played for the Dublin and Laois senior teams. Born in Abbeyleix, County Laois, McEvoy first played competitive hurling and Gaelic football in his youth. He arrived on the inter-county scene when he first linked up with the Laois senior team before later joining the Dublin senior team before returning to Laois. McEvoy was a regular member of the starting fifteen, and won one All-Ireland medal and two Leinster medals. He was an All-Ireland runner-up on one occasion. At club level McEvoy won several championship medals as a dual player with Abbeyleix. He also won a championship medal with the Thomas Davis club.}
\end{small}

\end{document}